# When the Feature Pool Goes Algorithmic: Extending Mufwene's Ecology of Language Evolution to LLM-Mediated Exposure

Kunmei Han[1]

[1] School of International Studies, Shenzhen University

**Abstract**

Mufwene's ecological model locates language evolution in competition among variants contributed by individual idiolects and in speakers' selection from linguistic material made available through interaction. Large language models (LLMs) complicate this architecture without requiring the locus of selection to move away from human speakers. This article argues that LLMs are best treated as distributional mediators: they aggregate language produced across human populations, transform its distribution through training and post-training, and redistribute model-specific outputs at scale. I call the resulting ecological process algorithmic reweighting of the speaker-accessible distribution: model mediation can alter the relative frequencies with which competing variants reach human selectors. Emerging evidence on model-specific linguistic profiles and lexical uptake is consistent with parts of this pathway, but does not establish inevitable convergence. Human social evaluation remains decisive: model-associated forms may diffuse and become conventionalized, become socially recognizable as 'AI-like' and subsequently avoided, or fail to diffuse in the first place. The proposal extends Mufwene's feature-pool ecology one step upstream of speaker selection and yields testable predictions about uptake, model-version effects, convergence, and social reversal.

## 1 The ecological problem posed by LLM-mediated exposure

Large language models (LLMs) occupy an unusual position in contemporary language ecologies. They are trained on human-produced language and, once deployed, their outputs can themselves become input to human communication (Yakura et al. 2026). Human language is therefore no longer only a source from which models learn: model-generated language can return to human interaction and potentially enter subsequent human production. The resulting feedback raises a question about language variation and change: what happens to competition among linguistic variants when widely deployed generative systems reshape the distribution of forms that speakers encounter?

Recent findings make this question empirically plausible. Matched human-LLM corpora show that models do not reproduce human linguistic distributions neutrally, but display recurrent lexical, grammatical, and rhetorical preferences (Reinhart et al. 2025). Other work has begun to connect model-side preferences with human usage. Yakura et al. (2026) quantify GPT-specific lexical preferences, identify corresponding post-release shifts in large-scale spontaneous speech, and experimentally show that brief exposure to variants supplied by an AI interlocutor can affect later lexical choice. Written scientific corpora likewise show sharp post-2022 increases in LLM-associated vocabulary and widespread LLM-modified text, although written data alone cannot distinguish direct AI assistance from human internalization (Juzek and Ward 2025; Liang et al. 2025). A separate literature raises a related issue about distributional diversity: repeated runs of a model may produce much narrower response distributions than human populations on some tasks (Park et al. 2024), while access to generative-AI ideas can make independently produced creative texts more similar even when individual performance improves (Doshi and Hauser 2024).

I propose that these findings, although originating in research on model behavior, human-AI interaction, and creativity, can be connected to a long-standing problem in evolutionary sociolinguistics: how variants become available for selection, how their ecological weights change, and how repeated individual selections accumulate into population-level language change. Mufwene's feature-pool model provides a particularly useful baseline because it explicitly locates selection in individual speakers and treats contact ecologies as conditions that structure which variants can be selected (Mufwene 2001, 2002).

The key claim of this article is that LLMs introduce an additional layer upstream of human selection. LLM development aggregates human-produced linguistic material and transforms output distributions through training and post-training; generation procedures further shape the distribution from which individual outputs are sampled (Ouyang et al. 2022; Holtzman et al. 2020; Reinhart et al. 2025). When such outputs are repeatedly redistributed to users, the relative frequencies with which linguistic variants are encountered may change (Yakura et al. 2026). I refer to the ecological consequence of this process as algorithmic reweighting of the speaker-accessible distribution.

This formulation preserves the human locus of selection while extending the ecology in which selection takes place. Model-specific weighting should predict later human uptake across many variants; abrupt model-version changes should sometimes be followed by lagged human changes; and social recognition of highly concentrated forms should be capable of reversing initially positive exposure effects. These are predictions of the present account of LLM use. Following Mufwene (2001, 2002), I use language evolution here to include changes in the structural and usage patterns of particular language varieties, not the phylogenetic emergence of the human language faculty.

## 2 Mufwene's theoretical baseline: feature pools, access, and speaker selection

Mufwene's ecological account begins from a population view of language. A communal language is not a single organism or a fully uniform grammar, but a construct extrapolated from the partially overlapping idiolects of individuals who recognize themselves as users of the same communicative code (Mufwene 2001, 2002). In the population-genetic analogy, the communal language is comparable to a species-level construct, whereas individual idiolects constitute the relevant individual-level realizations. Selection consequently operates directly at the level of idiolects, with its effects becoming visible at the communal level (Mufwene 2002).

In the feature-pool model, speakers contribute partially overlapping variants to a pool that may contain competing phonological, morphological, lexical, and grammatical forms, as well as alternative principles of usage. An individual does not acquire this communal pool wholesale. According to Mufwene (2002), language acquisition is better understood as a process of recreation: learners construct their own idiolects by selecting, recombining, and sometimes modifying features encountered in the utterances of different individuals with whom they interact. The linguistic options available to a learner are therefore shaped by the particular interactions through which the language is experienced. This constraint on access provides the point of departure for the extension proposed here.

The communal feature pool may be broad, but the linguistic material encountered by any individual speaker is socially and interactionally structured. Where a speaker acquires a language, with whom they interact, and the communication networks in which those interactions occur all form part of what Mufwene treats as the ecology of language (Mufwene 2001, 2002). In contact settings, variants associated with different dialects or languages may enter the same pool, but their population-level effects depend on whether speakers encounter and reproduce them sufficiently for idiolectal selections to spread beyond a small number of individuals. A feature may therefore be present in the broader communal ecology without being equally available, or equally influential, for all speakers.

Mufwene also distinguishes competition from agency. Linguistic variants do not literally compete as autonomous entities. Competition refers to the coexistence of alternatives that are unequally weighted, whereas selection is realized through the development and use of individual idiolects (Mufwene 2002). Some variants are favored and reproduced more often, while others remain marginal or fail to spread. Crucially, these outcomes arise through speaker-level selection under particular ecological conditions rather than through agency on the part of the variants themselves.

This distinction becomes important once LLMs are introduced into the analysis. A model may assign a relatively high generation probability to one form and a low probability to another, but such computational weighting is not equivalent to the speaker-level selection through which human idiolects are formed and communal language changes. The present extension therefore retains three commitments of Mufwene's framework: the individual idiolect remains the locus at which selection operates; selection depends on the linguistic alternatives made available through a speaker's ecology; and communal change emerges when recurrent idiolectal selections spread through a population. The new question arises one step upstream: what happens when the distribution of linguistic alternatives available to speakers is itself mediated by generative models?

## 3 What is an LLM in this ecology?

At a sufficiently abstract functional level, human speakers and LLMs share one relevant property: prior linguistic input conditions subsequent production. Usage-based theories have long emphasized that recurrent linguistic experience affects accessibility, routinization, and entrenchment in human speakers (Bybee 2010; Schmid 2020). Autoregressive language models likewise generate context-sensitive continuations from learned probability distributions; the realized output also depends on the generation procedure used to sample from those distributions (Holtzman et al. 2020). This functional similarity should not be taken to imply equivalent learning mechanisms or equivalent positions in the language ecology.

The ecological structures of the two systems differ fundamentally. A human idiolect is locally acquired and socially situated. Mufwene (2002) stresses that every speaker is exposed only to a subset of communal linguistic production because exposure is limited in space and time and depends on interaction with a subset of speakers. The resulting idiolect is therefore shaped by a particular interactional history and subsequently enters further human interaction as one situated source of linguistic variation.

An LLM distribution is better described as population-derived, algorithmically transformed, and redistributable at scale. Model development draws on large collections of human-produced text; post-training can systematically alter model behavior, and decoding choices affect the diversity and repetition of generated output (Ouyang et al. 2022; Holtzman et al. 2020). Corpus comparisons further show that the resulting outputs can exhibit model-specific lexical, grammatical, and rhetorical distributions that differ from matched human production (Reinhart et al. 2025).

*Human speaker: local exposure -> socially situated idiolect -> propagation through interaction*

*LLM: population-scale aggregation -> algorithmic transformation -> large-scale redistribution*

The similarity between speaker-mediated and LLM-mediated exposure is therefore functional, not ontological. Both can affect the linguistic material encountered by a human speaker. An LLM, however, should not be treated as simply another idiolect in Mufwene's feature pool. I use distributional mediator as a descriptive analytic label for a system derived from population-level linguistic production that transforms and redistributes a model-specific projection of that variation back into speakers' linguistic environments.

This distinction also clarifies what is distinctive about scale and network structure. Human linguistic exposure can of course be mass-mediated, but it normally draws on many distinct human authors and speakers. LLM-mediated exposure can instead present many otherwise unrelated users with individualized outputs sampled from a shared model distribution; Yakura et al. (2026) characterize this structure as parallel one-to-one interaction combined with concentration at the provider level. Users need not receive identical outputs for this structure to matter. The relevant claim is that many encounters can be probabilistically conditioned by the same underlying model and post-training regime. Table 1 summarizes the resulting analytic contrast between speaker-mediated and LLM-mediated exposure.

**Table 1:** Speaker-mediated exposure and LLM-mediated exposure in a Mufwenean feature-pool ecology. The speaker-mediated column synthesizes Mufwene (2001, 2002); the LLM-mediated column draws on Ouyang et al. (2022), Holtzman et al. (2020), Reinhart et al. (2025), and Yakura et al. (2026).

| Dimension | Speaker-mediated exposure | LLM-mediated exposure |
|---|---|---|
| Source of linguistic input | Socially situated interlocutors and human communicative artifacts | Human-produced linguistic material aggregated across large corpora and domains |
| Distribution presented to the user | Multiple locally or medially encountered human distributions | Model-specific output distribution sampled anew across interactions |
| Transformation of prior input | Human learning shaped by interactional history, accommodation, and social evaluation | Training and post-training alter model behavior; decoding procedures shape realized output |
| Network topology | Distributed across interpersonal, community, institutional, and media networks | Can be centralized around a shared model or model family used across many otherwise unrelated interactions |
| Scale of propagation | Ranges from local interaction to mass-mediated dissemination, usually involving many distinct human sources | A shared model-derived distribution can generate individualized outputs repeatedly across very large user populations |
| Temporal dynamics | Idiolectal and population changes diffuse through continuing interaction, although abrupt social shocks can occur | Model updates can abruptly change the distribution presented to many users |
| Role in selection | Human speakers encounter and select among variants | Model mediation reweights exposure; human speakers remain the locus of selection |
| Social meaning | Variants can index groups, identities, registers, and stances | Model-associated variants can additionally acquire indexical meanings such as 'AI-like' |

## 4 Extending the feature-pool model one step upstream

Mufwene's framework explains how individual speakers select among linguistic variants encountered through socially structured interaction (Mufwene 2001, 2002). The extension proposed here asks how the distribution presented to the speaker may itself be structured before selection takes place.

In the baseline account, the communal feature pool is not equally accessible to every speaker. Interlocutors, social networks, demographic configurations, communicative histories, and other ecological conditions determine the subset of variants that a speaker encounters (Mufwene 2001, 2002). More importantly for the present argument, variable exposure also implies unequal encounter frequencies among alternatives within that subset. The present account treats those relative encounter frequencies as part of the speaker-accessible distribution.

LLM deployment introduces an additional path through which this speaker-accessible distribution may be structured. Human-produced language enters model development; training and post-training transform model behavior; model-specific output is generated under particular decoding procedures; and that output can enter human linguistic environments either directly through human-LLM interaction or indirectly through AI-generated and AI-assisted content circulating in human communication (Ouyang et al. 2022; Holtzman et al. 2020; Liang et al. 2025; Yakura et al. 2026).

I refer to this process as algorithmic reweighting of the speaker-accessible distribution. The term identifies a process upstream of speaker selection: model mediation can alter the relative probabilities with which competing linguistic variants are encountered. If the communal feature pool contains competing variants f1,...,fn, the relevant input to an individual speaker is not the pool in its entirety but an ecology-conditioned distribution. Schematically:

$$P(fi \mid Es)$$

where Es denotes the speaker's broader ecology of exposure. An LLM-mediated component can be represented as an additional condition:

$$P(fi \mid Es, M)$$

where M denotes the model-mediated component of exposure. The notation is deliberately schematic. The theoretical claim is not that model-mediated input is necessarily narrower, more frequent, or more influential than other sources of input, but that P(fi | Es, M) may differ systematically from P(fi | Es) and can therefore alter the conditions under which subsequent human selection occurs. Figure 1 locates this proposed pathway relative to the simplified Mufwenean baseline: the added model-mediated path changes the speaker-accessible distribution while leaving speaker-level selection in place. The dashed return path marks a possible later training feedback loop and is not required for the core argument.

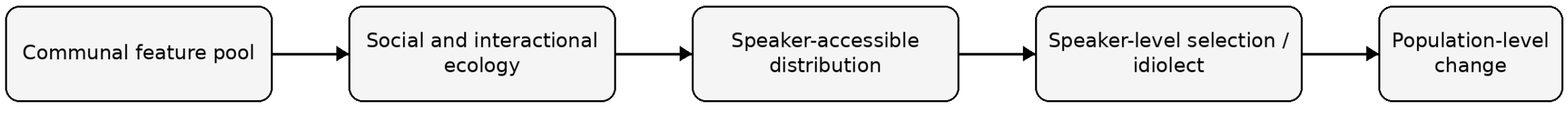

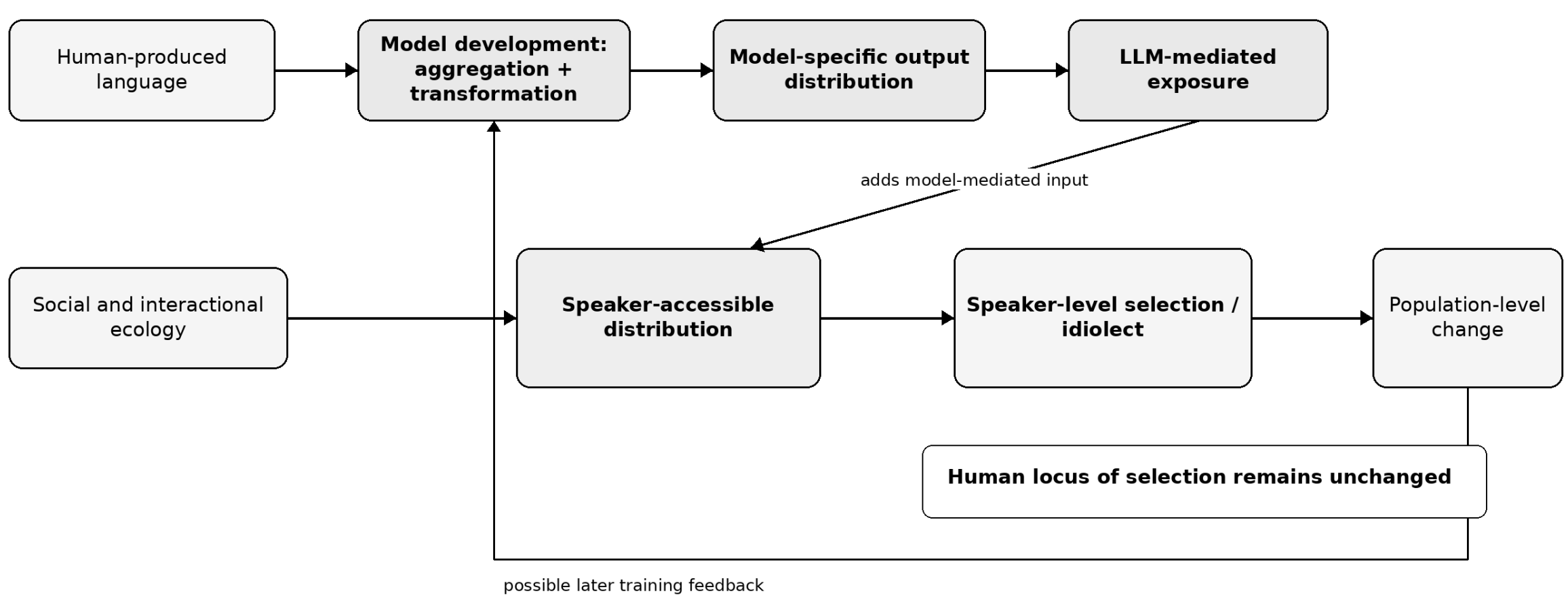

**Figure 1:** Extending the feature-pool ecology one step upstream. Panel A is a schematic reconstruction of the baseline developed from Mufwene (2001, 2002), in which social and interactional ecology structures the distribution of variants accessible to an individual speaker. Panel B presents the extension proposed in this article: population-derived linguistic material is transformed into a model-specific distribution and redistributed into the speaker's linguistic environment. The extension changes the distribution available for selection; it does not relocate the locus of selection away from human speakers.

Emerging empirical evidence provides preliminary support for selected links in this pathway. Reinhart et al. (2025) show that model-generated language can differ systematically from matched human production in lexical, grammatical, and rhetorical distributions. This establishes the first necessary condition for the proposed account: model mediation can transform human linguistic distributions.

Yakura et al. (2026) provide evidence for subsequent links. They quantify model-specific lexical preferences and show that more strongly ChatGPT-preferred words disproportionately increased in spontaneous human speech after the model's release. Their randomized experiment further shows that exposure to a lexical variant supplied by an AI interlocutor increased participants' later reuse of that variant after the immediate interaction had ended. In the present framework, these findings are consistent with a pathway from model-side reweighting to altered human exposure to subsequent speaker-level selection.

Taken together, these studies support selected links in the proposed pathway (i.e., model-side distributional divergence and lexical uptake. But they do not establish population-level convergence. That stronger claim requires evidence that competing alternatives become less diverse in human production, which motivates the next section.

# 5 Human selection after model-mediated exposure: convergence, enregisterment, and reversal

The distinction between diffusion and convergence is important. Algorithmic reweighting may increase the frequency of a model-preferred variant without reducing variation in the broader feature pool. Convergence requires a stronger pattern: competing alternatives must become less evenly distributed across human production.

Park et al. (2024) provide one reason to consider this possibility on the model side. On several psychological survey tasks, repeated GPT-3.5 runs produced zero or near-zero response variation where human samples were heterogeneous. Their dependent variables were judgments rather than linguistic variants, and some of the ’correct answer’ effects changed sharply when answer choices were reversed (Park et al. 2024). Their findings therefore motivate the possibility that model-generated distributions may sometimes be substantially more concentrated than corresponding human distributions. Doshi and Hauser (2024) provide a different form of human-side evidence: access to generative-AI ideas increased similarity among creative outputs produced by different participants even while improving some measures of individual performance. Taken together, these studies make distributional convergence a plausible empirical question about the consequence of LLM use.

Yet a Mufwenean account gives an equally strong reason not to assume that concentrated exposure produces inevitable convergence. Exposure constrains which alternatives can enter speaker-level competition and how frequently they are encountered, but it does not determine subsequent selection (Mufwene 2001, 2002). Speakers remain socially situated selectors, and linguistic forms can acquire new social meanings as they diffuse.

This possibility becomes especially important when model-associated forms become socially recognizable. Agha’s (2003) account of enregisterment describes how linguistic forms and repertoires become associated with socially recognizable styles and personae, while Eckert (2008) emphasizes that variants acquire indexical meanings through social use. An initially unremarkable model-preferred form may therefore change status as its association with AI becomes culturally legible.

A possible trajectory is:

*model concentration -> increased exposure -> human uptake -> social salience -> AI indexicality -> maintenance or avoidance*

Here the same concentration that favors diffusion can also create the conditions for resistance. A model-favored form may initially be an ordinary feature of the human pool whose only unusual property is its high probability under a model. If repeated redistribution raises exposure, the form may become more frequent in human production. As it becomes conspicuous, speakers may begin to recognize it as characteristic of machine-generated language. The form can then acquire new indexical meanings - for example, ’polished,’ ’generic,’ ’formulaic,’ ’professional,’ ’inauthentic,’ or simply ’AI-like’ - which can alter subsequent speaker-level selection (Agha 2003; Eckert 2008).

The resulting dynamics are therefore better understood as an interaction between model-side reweighting and human social evaluation. Model mediation can increase exposure to a form, while downstream social evaluation can stabilize, redirect, or reduce its reproduction. This does not require a separate evolutionary mechanism: it is still human selection, now operating on variants whose social meanings may themselves have been changed by model-mediated diffusion.

Recent trajectories of conspicuous AI-associated vocabulary are suggestive. Yakura et al. (2026) show that the strong early GPT preference for delve attenuates markedly across later model versions, while human use of the word rises after ChatGPT's release and subsequently reverts in several podcast categories. Because model-side supply changes at roughly the same historical moment that public awareness of delve as an AI-associated word increases, their observational trajectory does not isolate the cause of reversal. Written academic data show a related pattern: Geng and Trotta (2025) report declines in several publicly discussed ChatGPT-associated words after early 2024 and interpret these as compatible with user adaptation or editing, but such data cannot distinguish spontaneous human avoidance from deliberate modification of AI-assisted text. Reduced model-side supply and changed human-side evaluation are therefore analytically distinct mechanisms that should be separated whenever possible.

At least three outcomes are possible. Diffusion and conventionalization occur when increased exposure facilitates reuse while the form remains weakly marked. Boom and bust occurs when rapid diffusion increases social salience, AI-associated indexicality becomes negative or otherwise interactionally costly, and speakers reduce their use of the form. Ecological resistance occurs when existing register norms, identities, communicative goals, or entrenched alternatives outweigh the additional exposure and prevent substantial uptake in the first place.

# 6 Predictions and a research agenda

The proposed extension is useful only if it yields evidence that can discriminate among competing explanations. Table 2 summarizes four falsifiable predictions that follow most directly from the distributional-mediation account. Each concerns a relationship between model-side distributions, human exposure, and later human production.

**Table 2:** Falsifiable predictions derived from the distributional-mediation account. The expected patterns are author-derived predictions of the framework; the final column states observations that would weaken the corresponding.

| Prediction | Expected pattern | Critical evidence against the prediction |
|---|---|---|
| Preference-uptake correspondence | Across many competing forms, stronger and more stable model preferences predict greater later human uptake after baseline trajectory, topic, and register are controlled. | No graded association, or a reliable association in the opposite direction, in settings where model exposure can be independently established or credibly approximated. |
| Version tracking | A substantial version-specific change in model probability is followed by a lagged human change in the same direction, with larger effects where exposure is greater. | Human change consistently precedes the model shift, or exposed populations show no corresponding change while otherwise comparable conditions are met. |
| Concentration-convergence | When model output is narrower than the relevant human baseline, sustained exposure reduces between-speaker or between-text variation as well as shifting mean frequencies. | Favored forms increase but human variation remains stable or increases; this would support diffusion without supporting convergence. |
| Salience-reversal | Among similarly model-preferred forms, features that become socially recognizable as AI-associated are more likely to plateau, reverse, or become register-restricted. | AI salience has no moderating relationship with later use after model-side preference and exposure are controlled. |

The four predictions are related but test different links in the proposed pathway. Preference-uptake tests whether model-specific weighting is associated with later human change. Version tracking provides a stronger temporal test because changes in a deployed model can function as distributional shocks. Concentration-convergence asks whether model mediation changes variation itself. Salience-reversal tests the downstream role of human social evaluation.

Testing these predictions requires explicit measurement on both sides of the human-model relation. Model-side distributions should be estimated using repeated, controlled generations with model version, prompt, and generation settings recorded, as in parallel-corpus and model-comparison designs (Reinhart et al. 2025; Yakura et al. 2026). Human-side change requires longitudinal data in which time, register, and ideally speaker or author identity can be modeled. Written corpora present a special identification problem because apparent 'human' shifts may partly reflect AI-generated or AI-assisted text entering the corpus directly; large-scale estimates already indicate substantial LLM modification in scientific writing (Liang et al. 2025). Evidence for internalization is therefore stronger when model-associated changes persist in settings where the model is absent from the act of production, as in post-exposure production tasks or spontaneous speech (Yakura et al. 2026).

Causal leverage can come from several complementary designs. Parallel human-LLM corpora can identify model-specific feature distributions (Reinhart et al. 2025); longitudinal corpora can test whether those preferences predict later population change (Yakura et al. 2026); model-version shifts can provide quasi-experimental variation; and controlled exposure experiments can test persistence after the source of exposure is removed (Yakura et al. 2026). The analysis should move beyond lexical signatures to competing constructions, syntactic alternations, discourse markers, and information-structuring strategies, because the theoretical claim concerns distributions of linguistic alternatives.

A further priority is to specify ecological boundary conditions. The same model-side preference should not be expected to diffuse equally across communities, registers, platforms, or languages. In Mufwene's framework, uptake depends on the local ecology in which variants become available and are selected; demographic structure, interaction patterns, existing alternatives, and social evaluation can all alter population-level outcomes (Mufwene 2001, 2002). In the proposed extension, exposure intensity, entrenched competitors, social meaning, and communicative function should therefore be treated as moderators.

Finally, claims about cognitive or ideological convergence require an additional evidential step. Linguistic convergence alone cannot demonstrate convergence of thought. Such claims require independent dependent variables indexing conceptual organization, judgment, memory, or reasoning. Keeping this boundary explicit prevents a distributional account of linguistic change from being inflated into an unsupported theory of cognitive homogenization.

# 7 Conclusion

Mufwene's feature-pool model does not need to be abandoned in order to accommodate large language models. Its core commitments remain useful: communal language is grounded in populations of idiolects; competing variants are unequally weighted under particular ecological conditions; and speaker-level selection remains the locus through which human language changes. What LLM deployment changes is one ecological precondition of that selection: the distribution of linguistic alternatives that speakers encounter.

LLMs occupy this position because they combine population-scale aggregation, algorithmic transformation, and large-scale redistribution. They are therefore better treated as distributional mediators than as artificial speakers added to the feature pool. The proposed extension is correspondingly modest: it moves one analytical step upstream and asks how model-mediated exposure changes the speaker-accessible distribution before human selection occurs.

This extension also avoids technological determinism. Model-side reweighting can favor diffusion, but it does not determine the outcome. Human speakers can reproduce, conventionalize, reinterpret, resist, or avoid the variants they encounter. Concentrated model-associated forms may contribute to convergence under some conditions, while under others the same concentration may make a form salient enough to acquire AI-associated social meaning and trigger reversal.

The broader implication is that LLMs do not relocate the locus of language evolution from people to machines. They alter part of the ecology in which people select among variants. This creates a new empirical setting for testing long-standing claims about access, weighting, diffusion, and social selection, and potentially for observing recursive feedback when later human production becomes input to subsequent model generations. Whether such coupling produces convergence, diversification, resistance, or rapid cycles of adoption and obsolescence remains an empirical question.